\documentclass[letterpaper, 10 pt, conference]{ieeeconf}

\IEEEoverridecommandlockouts
\usepackage{balance}
\usepackage{amssymb}
\usepackage{graphics}
\usepackage{graphicx}
\usepackage{amsmath}
\usepackage{floatrow}
\usepackage{url}
\usepackage{cite}
\usepackage{textcomp}
\usepackage{gensymb}
\usepackage[ruled,vlined]{algorithm2e}
\usepackage{multirow}
\usepackage{tablefootnote}
\usepackage{siunitx}
\usepackage{tabularray}
\usepackage{multicol}

\title{\LARGE \bf
From Technical Metrics to User Perception: A User Study of a Multimodal
Human--Robot Interaction System for Object Detection and Grasping
}

\author{Jian Song, Tian Zi, Shen Guanting%
\thanks{The authors are with the Dalian University of Technology, Dalian, China. {\tt\footnotesize \{jian.song.robotics, zi.tian.robotics, shen.guanting.99\}@gmail.com}. The first author is the corresponding author.}%
}

\begin{document}

\maketitle
\thispagestyle{empty}
\pagestyle{empty}

\begin{abstract}
Improvements in the technical performance of human--robot interaction (HRI) systems do not automatically translate into differences that human users can detect during live interaction. This paper investigates whether a 15-percentage-point gain in end-to-end task success (from 75\% in a multimodal baseline system to 90\% in an improved configuration identified through a prior ablation study) is sufficient to produce consistent and measurable differences in user perception. The baseline system combines Whisper for speech recognition, Florence-2 for open-vocabulary object detection, LLaMA 3.1 for action extraction, and an interval Type-2 fuzzy logic controller for motion execution. The improved configuration replaces the perception and language modules with Grounding DINO + SAM and Qwen 3.5 9B, respectively, while retaining the same controller. A within-subject user study with 24 participants compared both systems on the same tabletop object-grasping task. After interacting with each configuration, participants rated perceived speed, reliability, and overall competence and fluency on a 7-point Likert scale. Results show that 17 out of 24 participants (70.83\%) preferred the improved system (exact binomial test, p = 0.043, h = 0.43), and all three perceptual constructs were rated significantly higher for the improved configuration after Holm correction, with large to very large effect sizes (p < 0.001). These findings confirm that the identified technical improvements are perceptible to users in direct interaction and underscore the importance of complementing benchmark evaluation with user-centred evidence when assessing robotic manipulation pipelines.
\end{abstract}



\section{Introduction}\label{sec:introduction}

Human--robot interaction (HRI) has moved steadily from rigid, task-specific automation toward
systems that can interpret human intent, adapt to changing contexts, and act in a way that
feels natural to non-expert users. This shift has been driven by applications in healthcare,
domestic assistance, education, logistics, and collaborative manufacturing, where robots are
expected to operate not only safely but also transparently and intuitively alongside
people~\cite{hu2011advanced,zhao2022human,he2017educational,kagami2006home}. Across these
settings, the accurate inference of human intention directly conditions system efficiency,
operational safety, and user satisfaction~\cite{huang2016anticipatory,lorentz2023pointing},
and has therefore become a core research problem in modern
HRI~\cite{dominguez2025human}. This requirement becomes especially
critical when the robot operates in safety-sensitive domains or in close physical proximity
to people: in robot-assisted surgery, misinterpretation of the operator's intent can have
severe consequences for patient safety~\cite{saeidi2019autonomous}, while in collaborative
transport and handover tasks, errors in intent inference can cause coordination failures
and physical harm~\cite{dominguez2023improving,dominguez2024exploring,dominguez2024force}.
When the robot must additionally respond to spoken instructions, visual context, and physical
constraints simultaneously, the design problem becomes inherently multimodal and must balance
perception, language understanding, and control under uncertainty.

Recent progress in foundation models has made this multimodal direction substantially more
practical. Large language models (LLMs) such as LLaMA and GPT-4 provide strong semantic
parsing and structured reasoning capabilities~\cite{touvron2023llama,achiam2023gpt}, while
vision--language models (VLMs) extend that ability to visual grounding by jointly reasoning
over images and text. This combination has motivated a growing body of work on robot
navigation, manipulation, and interactive assistance, including language-and-sketch
interfaces, multimodal navigation systems, and physically grounded vision--language reasoning
for manipulation tasks~\cite{zu2024language,zhang2024interactive,gao2024physically}. Beyond
task-level performance, multimodal interaction has also been shown to make robot behavior
more transparent and acceptable to end users, and to create opportunities for proactive robot
behaviors in which the system anticipates and responds to user needs before an explicit
command is issued~\cite{dominguez2025inference,dominguez2024anticipation}. At the same time,
the HRI literature has repeatedly cautioned that powerful generative models are not sufficient
on their own: without grounding in the physical world and deterministic downstream logic,
language models can be brittle, overconfident, or poorly aligned with robot
execution~\cite{atuhurra2024leveraging}. For this reason, many successful robotic systems
combine learned foundation models with explicit perception and control modules rather than
relying on a single end-to-end learned policy.

Our previous work followed this hybrid design philosophy and presented a multimodal
manipulation framework that tightly couples speech recognition, open-vocabulary visual
perception, natural language understanding, and adaptive motion control for a Dobot Magician
robotic arm~\cite{shen2026approach}. In that system, continuous wake-word activation was
handled by an Audio Spectrogram Transformer~\cite{gong2021ast}, spoken commands were
transcribed with Whisper~\cite{radford2023robust}, action intent was extracted by LLaMA~3.1,
object locations were inferred through Florence-2 in open-vocabulary
mode~\cite{xiao2024florence, dominguez2026ros}, and motion execution was governed by an interval Type-2 fuzzy
logic controller, chosen for its ability to explicitly model uncertainty in sensing and
actuation~\cite{mendel2002fuzzy,liang2000interval,hellmann2001fuzzy,hagras2004type}. The
resulting pipeline demonstrated the feasibility of spoken object manipulation in a controlled
tabletop setting and achieved an end-to-end success rate of 75\% across 60 trials, but it
also revealed clear bottlenecks. In particular, object perception and robot execution
accounted for most of the total latency, while action extraction and motion generation were
the main contributors to task failure. Those findings suggested that overall system
performance depended not only on the architecture, but also critically on the specific model
and controller choices made for each stage.

To understand those module-level effects more precisely, we then conducted a controlled
ablation study in which alternative language models, perception pipelines, and controllers
were compared under a common experimental protocol. That analysis showed that the perception
subsystem had the strongest influence on end-to-end success, that controller choice mainly
affected execution robustness under uncertainty, and that the language model primarily
influenced whether the generated action representation was valid and task-consistent. The
strongest combination among the tested modules improved end-to-end success substantially
beyond the original baseline, demonstrating that the technical gains were real. This result,
however, raises a more human-centered question: are those improvements large enough for
actual users to notice during live interaction?

This question motivates the present paper. Rather than proposing a new architecture, we
focus on whether people can perceive the difference between the original baseline system and
the strongest configuration identified in the ablation study. In other words, this work
shifts the evaluation emphasis from module-level technical benchmarking to user-level
perception of performance differences. This perspective matters because improvements in
success rate or latency do not automatically translate into perceptible differences for
human observers, especially when the task is short, the scene is constrained, and the robot
behavior is already relatively stable in both conditions~\cite{dominguez2025inference}. A user study can therefore reveal
whether the technical improvements identified in previous work are meaningful from an
interaction standpoint, not only from a benchmark standpoint.

The contributions of this paper are threefold. First, we introduce a focused within-subject
user study that directly compares the baseline multimodal manipulation system with the
best-performing configuration selected from the ablation study. Second, we analyze whether
participants can distinguish the two systems in terms of perceived speed, reliability,
smoothness, and overall quality of execution, and whether their subjective judgments align
with the objective performance evidence. Third, we discuss the implications of the findings
for how HRI systems should be evaluated, arguing that technical metrics and user perception
are complementary dimensions that should be assessed jointly when validating improvements to
robotic manipulation pipelines. The remainder of the paper is organized as follows: the next
section summarizes the baseline and improved systems, followed by the experimental design,
results, discussion, and conclusions.


\section{Baseline and Improved System Recap}
\label{sec:baseline}

\begin{figure}[t]
	\centering
	\includegraphics[width=0.98\textwidth]{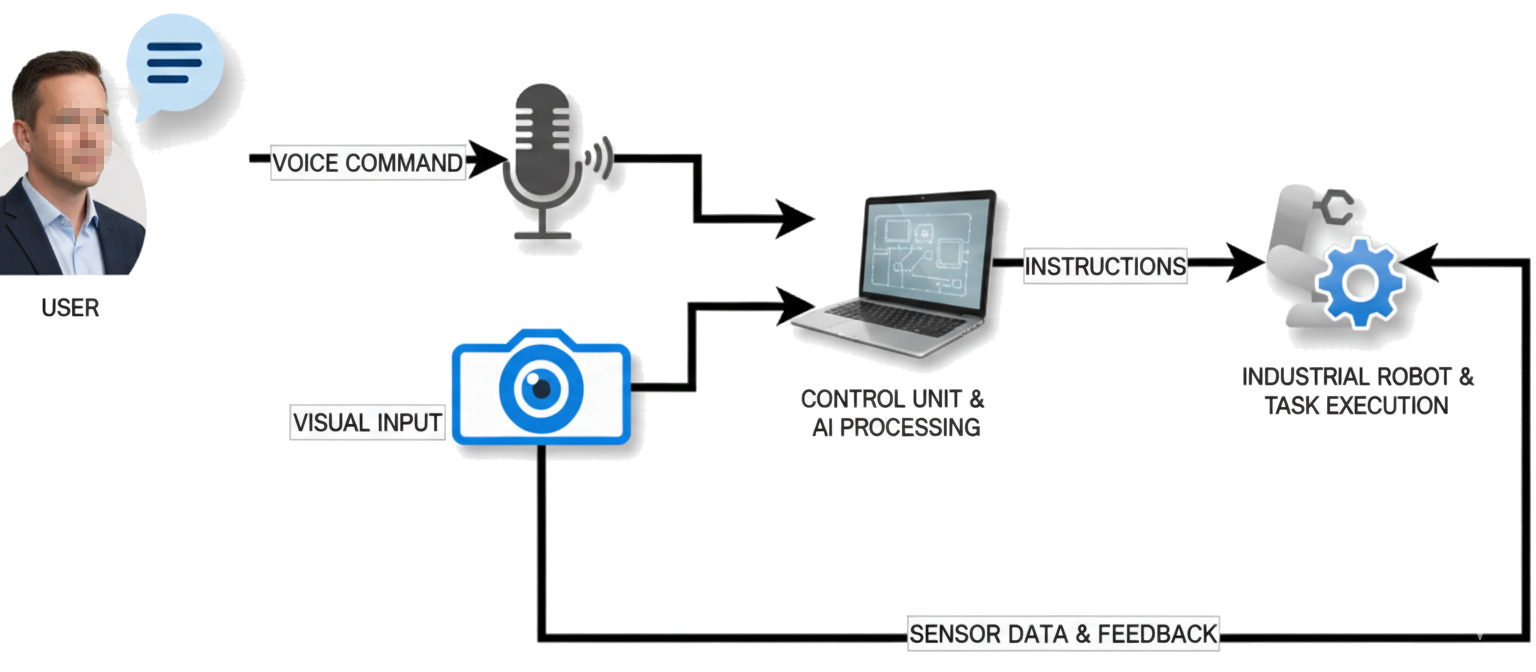}
	\caption{{\bf Information flow of the complete system for both pipelines.} Hardware elements and data flow of the robotic manipulation pipeline. Figure from~\cite{shen2026approach}.}
	\label{fig:Block-diagram}
\end{figure}

The user study compares two configurations of the same tabletop manipulation pipeline. Both systems share the same robot, camera, audio front end, computation platform, workspace geometry, and command protocol, so the comparison isolates the effect of the perception, language grounding, and control modules rather than physical setup differences. To make the comparison easier to follow, Fig.~\ref{fig:Block-diagram} summarizes the common task setting, and Table~\ref{tab:system_comparison} lists the module-level differences between the baseline and improved configurations.


\begin{table*}[t]
\centering
\caption{Concise comparison of the two systems evaluated in the user study.}
\label{tab:system_comparison}
\renewcommand{\arraystretch}{1.15}
\begin{tabular}{p{2.8cm}p{5.2cm}p{7.8cm}}
\hline
\textbf{Module} & \textbf{Baseline system} & \textbf{Improved system} \\
\hline
Language grounding & LLaMA 3.1 7B interprets the transcribed command and converts it into a structured robot action list. & Qwen 3.5 9B performs the same action-extraction task and was selected because it produced the strongest language-stage results in the ablation study. \\
Visual grounding & Florence-2 performs open-vocabulary object detection on the tabletop scene. & Grounding DINO + SAM provides the strongest visual grounding among the tested perception pipelines and serves as the improved perception module. \\
Motion control & Interval Type-2 fuzzy logic control executes the robot motion and handles uncertainty in the grasping step. & The same IT2FLS controller is retained because it remained the best controller choice in the ablation study. \\
Reported end-to-end performance & The baseline article reported 75\% overall task success across 60 trials. & The best configuration in the ablation study reached 90\% success in the combined comparison stage. \\
\hline
\end{tabular}
\end{table*}

\subsection{Hardware Platform and Fixed Infrastructure}

The comparison was conducted on exactly the same hardware platform described in the baseline article \cite{shen2026approach}. The robotic arm is a Dobot Magician equipped with a suction-based end-effector, visual sensing is provided by an Intel RealSense D435i RGB-D camera, and spoken commands are captured through Samsung Buds2 wireless earbuds. All inference and control routines run locally on an Intel Core i9-14900HX laptop with an NVIDIA RTX 4070 GPU. Keeping the hardware unchanged is important for the user study because it ensures that any perceived difference between the two systems is caused by the software stack rather than by different sensors, compute resources, or robot mechanics.

The experimental workspace is also held constant. The camera viewpoint, robot base pose, tabletop geometry, lighting conditions, and calibration procedure are preserved across both configurations. In addition, the same graphical user interface, logging logic, and safety checks are used in both conditions so that the only substantive differences are the modules under comparison. This fixed infrastructure allows the user study to focus on the perceptual consequences of improved language grounding and perception quality while leaving the rest of the pipeline unchanged.

\subsection{Task Definition}

The task is spoken-command object grasping in a controlled tabletop environment. Participants issue short natural-language requests that ask the robot to pick up, move, or spatially relate fruit objects placed on the table. The command space follows the same structure as in the earlier papers: simple imperatives are used for single-object pickup tasks, while some commands involve relative placement expressions such as left of, right of, next to, or in front of another object. This design keeps the language-to-action mapping transparent and allows the user study to reuse the same task family that was already evaluated technically in the two previous articles.

The object vocabulary and scene layout are also unchanged. Printed fruit images are used instead of physical fruit so that the robot can be evaluated under a consistent appearance and geometry across trials. A trial is considered successful only when the system correctly interprets the command, grounds the intended target object, and completes the manipulation without control failure or selection error. This binary success criterion is the same one used in the earlier studies, which makes the comparison between the baseline and improved systems directly interpretable.

\subsection{Robotic System}

At a high level, both systems implement the same multimodal pipeline. A spoken command is first captured by the audio front end, then transcribed into text, then transformed into a structured action description, then grounded in the visual scene, and finally executed by the robot controller. The difference between the two conditions lies not in the overall architecture, but in which language model and perception backend are used to realize the same processing stages. Because the user study focuses on whether people can perceive performance differences, this section emphasizes the components that changed and the components that remained fixed.

\subsubsection{Original System}

The baseline system described in the first article uses a multimodal interaction stack built around wake-word detection, speech-to-text transcription, semantic action extraction, visual object detection, and fuzzy control. Wake-word recognition is handled by an Audio Spectrogram Transformer, which monitors the incoming audio stream until the activation phrase is detected. Once activated, Whisper converts the spoken command into text. The resulting transcription is passed to LLaMA 3.1 7B, which extracts the intended manipulation action and converts it into a structured command representation. For visual grounding, Florence-2 is used in open-vocabulary mode to localize the requested object in the tabletop scene. The robot’s end-effector position is tracked using an ArUco marker, and motion execution is governed by an interval Type-2 fuzzy logic controller.

The baseline article showed that this configuration is feasible but still limited in reliability and speed \cite{shen2026approach}. Over 60 trials, the full pipeline achieved an overall task success rate of 75\%, while object detection and robot execution accounted for most of the total latency. The error analysis also showed that action extraction and robot motion were the largest contributors to failure. These findings are important for the present study because they establish the baseline against which the improved configuration is judged by users.

\subsection{Improved System}

The improved configuration is the strongest combination identified in the ablation study. In this condition, the language model used for action extraction is replaced by Qwen 3.5 9B, and the visual grounding module is replaced by Grounding DINO + SAM. The interval Type-2 fuzzy logic controller is retained, because it remained the best-performing controller in the ablation analysis and already provided the most stable motion execution among the tested alternatives. In short, the improved system keeps the same overall pipeline and hardware but uses stronger models in the stages that contributed most to the baseline system’s variability.

The ablation study showed that the perception module had the largest effect on end-to-end success, the language model mainly affected whether commands were parsed into valid actions, and the controller mainly affected robustness during execution. The best combined configuration therefore paired the strongest perception backend with the most reliable control strategy and a language model that produced the most consistent structured outputs. This combination achieved the highest success rate in the combined comparison stage, reaching 90\% over the tested trials. For the user study, this configuration is the most appropriate improved condition because it represents the clearest technical upgrade over the baseline while remaining faithful to the same task and experimental setting.

The purpose of comparing these two systems is not simply to confirm that one is objectively better than the other. Rather, it is to determine whether the improvements identified in the technical ablation study are also noticeable to human observers. The remainder of the paper therefore treats the baseline and improved systems as two versions of the same robot behavior pipeline, differing only in the internal models used for grounding and action selection.

\section{Experimental Design}
\label{sec:design}

The user study was designed to determine whether the technical improvements identified in the ablation study are also perceptible to human users during interaction with the robot. Accordingly, the study compares the baseline system introduced in the first article with the improved configuration selected in the second article, while keeping the task, robot platform, workspace, and command format unchanged from the previously reported experiments \cite{shen2026approach}.

A within-subject design was adopted so that every volunteer evaluated both systems under the same task conditions. Each participant completed two runs of the grasping task with one system, answered the questionnaire for that system, and then repeated the same procedure with the other system. In total, each volunteer therefore completed four live runs: two with the baseline system and two with the improved system. The order of the systems was randomized across participants to reduce learning, fatigue, and expectation effects. The same questionnaire was used after both systems, which made the two experiences directly comparable. All trials were carried out as live interactions with the physical robot rather than as video-based evaluations.

The task itself was the same tabletop object-manipulation task described in the previous articles. Participants issued short natural-language commands referring to fruit objects in the scene, and the robot was expected to identify the correct target and execute the requested grasping or relative-placement action. Because the task definition, scene layout, camera position, calibration procedure, and object vocabulary were preserved, any difference reported by the participants can be interpreted as a response to the software configuration rather than to changes in the experimental setup.

A total of 24 volunteers participated in the study. Only adults of legal age who were in full possession of their faculties and able to understand the instructions were accepted. Each session lasted approximately 30 minutes, including a short explanation of how the robot works, the purpose of the study, the two runs for each system, and completion of the questionnaires. No participant was asked to evaluate the internal implementation of the system; instead, the focus was on the subjective experience of using each version of the robot in the same task.

To minimize bias, the two systems were presented neutrally during the session, for example as System A and System B. Participants were told only that they would compare two versions of the same object-grasping pipeline and that both versions would be tested under identical conditions. The system labels were randomized across participants, and the final preference question was collected only after both systems had been experienced. This procedure allowed the study to capture relative user judgments without revealing which configuration corresponded to the baseline article and which corresponded to the improved system from the ablation study.

\subsection{User Study Research Questions and Hypotheses}

The user study addresses three research questions. First, can participants reliably distinguish the baseline system from the improved system after using both of them in the same task? Second, do participants perceive differences in speed, reliability, fluency, and overall competence between the two systems? Third, does the system preferred by participants align with the configuration that achieved the strongest objective performance in the technical evaluation?

These questions are grounded in the results of the previous two articles. The baseline study established that the original multimodal pipeline was feasible but still limited by object perception and action execution, with an overall end-to-end success rate of 75\% \cite{shen2026approach}. The ablation study then showed that the strongest configuration was obtained by combining Qwen 3.5 9B for action extraction, Grounding DINO + SAM for perception, and IT2FLS for control, yielding the highest success rate among the tested combinations. Because that improved configuration outperformed the baseline in the technical study, we expected users to notice at least some of those gains in everyday interaction.

From these observations, we formulated three hypotheses. H1 predicts that participants will rate the improved system as faster or more responsive than the baseline. This expectation is justified by the ablation study, which showed that the improved configuration can reduce delays in the perception and control stages relative to the original system. H2 predicts that participants will rate the improved system as more reliable and more likely to behave correctly during the task. This expectation follows directly from the higher task success reported for the best-performing combination in the second article. H3 predicts that participants will judge the improved system as smoother and more competent overall and will prefer it in the final choice question. This hypothesis reflects the combination of better perception, more robust control, and stronger language grounding observed in the ablation study.

The questionnaire was therefore constructed around these three constructs: perceived speed, perceived reliability, and overall competence or fluency. For each construct, participants answered two statements on a 7-point Likert scale, with 1 indicating strong disagreement and 7 indicating strong agreement. The same questionnaire was completed after evaluating each system so that ratings could be compared directly within participant. A final forced-choice question was used to determine which system the participant preferred overall.

\subsection{Statistical Analysis Procedure}

The dependent variables collected in the study are the questionnaire ratings, the composite scores derived from each pair of statements, and the final system-choice response. Because the design is paired and each participant evaluated both systems, the primary comparisons are within-subject comparisons between the baseline and improved configurations. The questionnaire uses 7-point Likert items, so the data are ordinal at the item level and should be analyzed accordingly.

For each construct, the two item scores will first be checked for internal consistency, and the resulting construct score will be obtained by averaging the two responses. The normality of the paired difference scores will then be assessed using the Shapiro--Wilk test. Because the data are paired and the response scale is ordinal, the default inferential test will be the Wilcoxon signed-rank test. If a construct score is approximately normal, a paired two-tailed t-test may be reported as a sensitivity analysis. In either case, effect sizes will be reported together with the test statistics; Cohen's $d_z$ will be used for paired t-tests, and the rank-biserial correlation or the standardized Wilcoxon effect size will be used for nonparametric comparisons.

The final preference question will be analyzed with an exact binomial test against a 0.5 null proportion, since each participant can select only one of the two systems. If needed, the influence of presentation order can be checked with an additional linear mixed-effects model using participant as a random intercept and system order as a fixed effect, but the main inferential results will be based on the within-subject comparisons described above. All hypothesis tests will be two-tailed with a significance threshold of $\alpha = 0.05$. When multiple construct-level comparisons are reported together, Holm correction will be applied to control the family-wise error rate.

In addition to the quantitative tests, the results section will also report descriptive summaries of the Likert responses, including central tendency and dispersion, so that the statistical conclusions can be interpreted together with the actual distribution of user judgments.

\section{Results}
\label{sec:results}

This section reports the outcomes of the user study following the analysis procedure specified in
Section~\ref{sec:design}. Results are organized by research question and by the corresponding
hypotheses H1--H3 introduced in Section~\ref{sec:design}. Unless otherwise stated, the significance
threshold is $\alpha = 0.05$; $p$-values for the three construct comparisons are Holm-corrected; and
effect sizes accompany every inferential test. A compact summary of all inferential statistics is
provided in Table~\ref{tab:results_summary}.

\subsection{Participant Demographics}
\label{sec:demographics}

Twenty-four adults voluntarily completed the study. The sample had a mean age of 24.6 years
($SD = 4.36$), and 37.5\% of participants identified as female (9 out of 24). All participants were
of legal age and in full possession of their faculties, as required by the inclusion criteria
described in Section~\ref{sec:design}. No participant was excluded from the analysis.

\subsection{Research Question 1: Ability to Distinguish the Two Systems}
\label{sec:rq1}

RQ1 asked whether participants could reliably identify a performance difference between the baseline
and the improved system after interacting with both under the same task conditions. The forced-choice
preference item collected at the end of each session was used to address this question.

Seventeen out of 24 participants (70.83\%) indicated an overall preference for the improved system.
An exact two-tailed binomial test against a null proportion of 0.50 yielded $p = 0.043$, with a
95\% Clopper--Pearson confidence interval for the true proportion of $[0.49,\ 0.87]$. The
corresponding effect size was moderate (Cohen's $h = 0.43$).

These results indicate that participants were able to distinguish the two configurations above chance
level at the specified significance level, supporting the directional component of H3. At the same
time, the moderate effect size and the wide confidence interval are consistent with the nature of the
difference being evaluated: the improved configuration outperformed the baseline by approximately 15
percentage points in end-to-end task success, but both systems share the same platform, user
interface, and command format. Under these conditions, a majority preference that is significant at
$\alpha = 0.05$ but not dramatic is a plausible and informative outcome, since perceptual differences
that arise primarily from reduced failure rates may not be salient during every individual trial.

\subsection{Research Question 2: Perceived Performance Differences}
\label{sec:rq2}

RQ2 addressed whether participants perceived differences in speed, reliability, and overall
competence and fluency between the two systems. For each construct, the composite score was obtained
by averaging the two corresponding Likert items from the questionnaire (see Appendix~\ref{app:questionnaire}).
Because the Shapiro--Wilk test applied to the paired difference scores did not reject normality for
any construct (all $W > 0.93$, $p > 0.05$), the paired two-tailed $t$-test was used as the primary
inferential test, consistent with the pre-specified sensitivity analysis described in
Section~\ref{sec:design}. All reported $p$-values for this research question have been Holm-corrected
for the family of three simultaneous comparisons. Effect sizes are expressed as Cohen's $d_z$, defined
as the ratio of the mean paired difference to the standard deviation of the paired differences.

\subsubsection{Construct 1: Perceived Speed (H1)}

Participants rated the improved system as substantially more responsive than the baseline
(baseline: $M = 4.61$, $SD = 0.55$; improved: $M = 5.69$, $SD = 0.61$). The mean paired difference
was $\Delta = 1.08$ scale points (95\% CI: $[0.83,\ 1.33]$), corresponding to a very large effect
size ($d_z = 1.85$). The paired $t$-test was highly significant ($t(23) = 9.09$,
$p_{\mathrm{adj}} < 0.001$), confirming H1. The improved system's mean rating of 5.69 out of 7
positions participants firmly above the midpoint on the responsiveness dimension, whereas the
baseline mean of 4.61 sits close to the neutral point, suggesting that the original configuration's
long average completion time (35.37 s across 60 trials in the baseline study) was perceptible to
users as a source of delay.

\subsubsection{Construct 2: Perceived Reliability (H2)}

The improved system was also rated as significantly more reliable than the baseline
(baseline: $M = 3.95$, $SD = 0.63$; improved: $M = 4.89$, $SD = 0.59$). The mean paired
difference was $\Delta = 0.94$ scale points (95\% CI: $[0.68,\ 1.20]$), with a large effect size
($d_z = 1.54$). The paired $t$-test was again highly significant ($t(23) = 7.54$,
$p_{\mathrm{adj}} < 0.001$), supporting H2. Notably, the baseline mean of 3.95 falls just below the
scale midpoint, indicating that participants on average perceived the original system as slightly
unreliable. The improved system's mean of 4.89 is above the midpoint but still well short of the
scale ceiling, which is consistent with the ablation study observation that the best configuration
reached 90\% end-to-end success but still exhibited occasional task failures. The reliability
construct therefore captures a meaningful subjective difference while also reflecting the fact that
neither version was perceived as fully dependable.

\subsubsection{Construct 3: Perceived Competence and Fluency (H3)}

The largest perceptual difference was observed for overall perceived competence and fluency
(baseline: $M = 3.76$, $SD = 0.59$; improved: $M = 5.01$, $SD = 0.64$). The mean paired difference
was $\Delta = 1.25$ scale points (95\% CI: $[0.99,\ 1.51]$), and the associated effect size
($d_z = 2.03$) is the largest in the study. The paired $t$-test was highly significant
($t(23) = 9.93$, $p_{\mathrm{adj}} < 0.001$), supporting the subjective dimension of H3. The
baseline mean of 3.76 falls below the midpoint, while the improved system's mean of 5.01 is clearly
above it. This construct, which integrates judgments of movement smoothness and overall task
capability, appears to be the dimension most sensitive to the combined improvements in visual
grounding, action extraction, and control robustness that distinguish the two configurations.
Together with the preference result in Section~\ref{sec:rq1}, this construct provides the strongest
evidence that users experienced the improved system as qualitatively more capable.

\begin{table*}[t]
\centering
\caption{Summary of inferential statistics for Research Questions 1 and 2.}
\label{tab:results_summary}
\renewcommand{\arraystretch}{1.2}
\begin{tabular}{p{5.9cm}p{2.0cm}p{1.1cm}p{2.0cm}p{1.7cm}}
\hline
\textbf{Measure} & \textbf{Test statistic} & \textbf{$p$} & \textbf{Effect size} & \textbf{95\%~CI} \\
\hline
Preference (17/24, 70.8\%) & Binomial & 0.043 & $h = 0.43$ & $[0.49, 0.87]$\tablefootnote{Clopper--Pearson interval for the true proportion.} \\
Perceived Speed ($\Delta = 1.08$) & $t(23) = 9.09$ & ${<}0.001$ & $d_z = 1.85$ & $[0.83, 1.33]$ \\
Perceived Reliability ($\Delta = 0.94$) & $t(23) = 7.54$ & ${<}0.001$ & $d_z = 1.54$ & $[0.68, 1.20]$ \\
Perceived Competence ($\Delta = 1.25$) & $t(23) = 9.93$ & ${<}0.001$ & $d_z = 2.03$ & $[0.99, 1.51]$ \\
\hline
\end{tabular}
\end{table*}

\subsection{Research Question 3: Relationship Between Objective and Subjective Performance}
\label{sec:rq3}

RQ3 asked whether the system preferred by participants was the same one that achieved the stronger
objective performance in the technical evaluation. The answer is affirmative across all measured
dimensions. On the preference item, 17 out of 24 participants chose the improved system, which is
also the configuration that reached the highest end-to-end task success rate (90\%) in the ablation
study compared with the 75\% achieved by the baseline. On the construct scales, the improved system
received higher ratings on all three dimensions, and in every case the differences were highly
significant after Holm correction, with large to very large effect sizes.

This alignment between objective and subjective measures is noteworthy because it is not guaranteed
in HRI evaluation. Technical improvements can sometimes be too small, too infrequent, or expressed
in dimensions that are not salient to non-expert users during live interaction, leaving no detectable
trace in subjective experience. In the present study, however, the 15-percentage-point gain in task
success appears to have been sufficient to generate consistent and measurable perceptual differences
across all three constructs. The most pronounced subjective gain was observed for perceived
competence and fluency ($d_z = 2.03$), the construct that captures participants' holistic impression
of the robot's overall behavior, suggesting that users were able to integrate local cues---fewer
selection errors, more stable motion, and reduced waiting time---into a coherent global judgment
of the system's capability.

The reliability construct, although significantly improved ($d_z = 1.54$), showed the smallest
absolute effect among the three and left the improved system's mean rating close to the midpoint of
the scale. This pattern is consistent with the ablation study finding that even the best-performing
configuration still exhibited occasional task failures and that the controller, though more robust
than the alternatives, does not eliminate execution errors entirely. From an HRI evaluation
perspective, this dissociation is informative: participants noticed and valued the improvement in
reliability but also detected that the system had not yet reached the level of dependability they
would expect from a fully trustworthy robotic assistant. That residual perception of unreliability,
even after significant technical improvements, underscores the importance of including user-study
evidence when assessing readiness for real-world deployment.

\subsection{Summary of Findings}
\label{sec:summary}

All three pre-registered hypotheses were supported by the data. H1---predicting higher perceived
speed ratings for the improved system---was confirmed by a very large effect ($d_z = 1.85$,
$p_{\mathrm{adj}} < 0.001$). H2---predicting higher reliability ratings---was confirmed with a large
effect ($d_z = 1.54$, $p_{\mathrm{adj}} < 0.001$), accompanied by the observation that neither system
reached clearly high absolute ratings on this construct. H3---predicting a smoother and more
competent overall impression, and a user preference for the improved configuration---was confirmed
by both the largest construct-level effect in the study ($d_z = 2.03$, $p_{\mathrm{adj}} < 0.001$)
and the preference result ($p = 0.043$, $h = 0.43$, 95\% CI $[0.49, 0.87]$).

Taken together, these findings demonstrate that the technical gains documented in the ablation study
are not only quantitative improvements on benchmark metrics but are also perceptible to users in
direct live interaction. This has an important practical implication: evaluating multimodal
manipulation systems solely through end-to-end task success rate or completion time is insufficient,
because those technical metrics do not capture how improvements translate into human experience. At
the same time, the moderate magnitude of the preference effect ($h = 0.43$) and the near-midpoint
reliability ratings indicate that the improved configuration, while clearly superior to the baseline
under the tested conditions, is still short of the level of seamlessness and consistency that users
would require from a system ready for deployment in uncontrolled real-world environments.

\section{Conclusion}
\label{sec:conclusion}

This work completes a three-paper research trajectory on multimodal spoken-command object
manipulation. The first paper introduced a unified HRI pipeline that combined Whisper for
speech recognition, Florence-2 for open-vocabulary visual grounding, LLaMA~3.1 for semantic
action extraction, and an interval Type-2 fuzzy logic controller for motion execution,
achieving an end-to-end task success rate of 75\% over 60 live trials~\cite{shen2026approach}.
The second paper conducted a controlled ablation study across alternative language models,
perception pipelines, and controllers, demonstrating that the strongest combination---Qwen~3.5~9B,
Grounding~DINO~+~SAM, and IT2FLS---reached 90\% success and that the perception module was the
single most influential factor in end-to-end performance. The present paper addressed the
question left open by that technical evidence: whether the documented improvements are also
detectable by human users interacting with the physical robot.

The user study showed that they are. Seventeen out of 24 participants (70.83\%) preferred
the improved configuration over the baseline in a forced-choice question, a result that is
statistically significant at $\alpha = 0.05$ ($p = 0.043$, Cohen's $h = 0.43$). On all three
perceptual constructs measured by the questionnaire---perceived speed, perceived reliability,
and overall competence and fluency---the improved system received substantially higher ratings,
with Holm-corrected $p$-values below 0.001 and effect sizes ranging from large ($d_z = 1.54$
for reliability) to very large ($d_z = 2.03$ for competence and fluency). The pattern of
subjective judgments aligned closely with the objective technical evidence: the system that
performed best in the ablation study was also the system that users found faster, more
reliable, and more coherent in motion. This alignment is meaningful because it is not
guaranteed---technical improvements that are small in magnitude, infrequent in occurrence,
or expressed in dimensions that non-expert observers cannot readily interpret often leave no
trace in user experience. In the present case, a 15-percentage-point gain in end-to-end
success proved sufficient to cross the threshold of perceptibility.

These findings carry a methodological implication for HRI evaluation more broadly. Technical
metrics such as task success rate, per-stage accuracy, and completion time are necessary but
not sufficient for assessing the quality of an interaction system from the user's point of
view. The present study provides evidence that subjective ratings and objective performance
can align, and that user-centred evaluation should therefore be included as a routine step
when validating improvements identified through ablation studies or benchmark comparisons.
At the same time, the moderate preference effect size and the near-midpoint absolute
reliability ratings for the improved system both suggest that the current implementation
is still some distance from the level of seamlessness and dependability that users would
require in unconstrained real-world scenarios. This underscores the need to treat the
alignment between benchmarks and perception as a question to be answered empirically rather
than as an assumption.

Several limitations define the scope of these conclusions and point to directions for future
research. The sample size of 24 participants is modest, and the study was conducted in a
controlled laboratory environment with a constrained object vocabulary, printed stimuli, and
a brief per-condition protocol. Consequently, the conclusions should be interpreted as
evidence of perceptibility within this specific task and setting, rather than as a general
characterisation of how users respond to improvements of this magnitude in richer or more
naturalistic environments. Future work should address these constraints by recruiting larger
and more diverse participant populations across different levels of prior robotics experience,
extending the task to longer sessions and more complex instruction sets that include
multi-object manipulation and dynamic scene changes, and adopting a broader evaluation
framework that captures trust, perceived safety, cognitive workload, and long-term acceptance
in addition to task-performance perception. Longitudinal designs would be particularly
valuable for understanding whether the perceptual advantages of the improved configuration
are maintained as users become familiar with both systems and as the novelty effect
diminishes. Replacing the printed fruit images with physically varied real-world objects
would also increase ecological validity and provide a more stringent test of the visual
grounding component, which the ablation study identified as the primary determinant of
end-to-end performance.


\appendix
\section{User Questionnaire}
\label{app:questionnaire}

The same questionnaire was administered after participants completed the two runs corresponding to each system. During the session, the systems were labeled neutrally as System A and System B. Each statement below was rated on a 7-point Likert scale, where 1 means strongly disagree and 7 means strongly agree.

\subsection*{Construct 1: Perceived speed}

\begin{enumerate}
    \item The system responded quickly to my commands.
    \item The system completed the task without unnecessary waiting or delay.
\end{enumerate}

\subsection*{Construct 2: Perceived reliability}

\begin{enumerate}
    \setcounter{enumi}{2}
    \item The system selected the correct object reliably.
    \item The system behaved consistently during the task.
\end{enumerate}

\subsection*{Construct 3: Overall competence and fluency}

\begin{enumerate}
    \setcounter{enumi}{4}
    \item The system's movements looked smooth and well coordinated.
    \item Overall, this system felt more capable for this task.
\end{enumerate}

\subsection*{Final choice question}

\begin{enumerate}
    \setcounter{enumi}{6}
    \item Which system did you prefer overall for this task?\\
    \hspace*{1em}$\square$ System A \hspace*{2em}$\square$ System B
\end{enumerate}

\bibliographystyle{IEEEtran}
\bibliography{IEEEabrv,./bib.bib}

\end{document}